\documentclass[journal]{IEEEtran}

\usepackage[ruled,boxed, linesnumbered, vlined, nofillcomment]{algorithm2e}
\usepackage{subfigure,amssymb,amsmath,graphicx,epsfig,multirow,url}
\usepackage{setspace,color}
\usepackage{comment}
\usepackage{enumitem}

\usepackage{algorithmic}
\usepackage{amsfonts}

\usepackage{booktabs}

\ifCLASSINFOpdf
\else
\fi

\begin{document}
\title{Half a Dozen Real-World Applications of Evolutionary Multitasking, and More}
\author{~Abhishek~Gupta,
        ~Lei~Zhou*,~Yew-Soon~Ong,~Zefeng~Chen,~Yaqing~Hou
%\thanks{Abhishek Gupta is with the Agency for Science, Technology and Research (A*STAR), Singapore. He also holds a secondary affiliation with the School of Computer Science and Engineering (SCSE), Nanyang Technological University (NTU), Singapore. Email: abhishek\_gupta@simtech.a-star.edu.sg.}
%\thanks{Yew-Soon Ong is with A*STAR, Singapore. He is also with the SCSE, NTU. Email: asysong@ntu.edu.sg}
%\thanks{Lei Zhou and Zefeng Chen are with the SCSE, NTU, Singapore. Email: \{lei.zhou, zefeng.chen\}@ntu.edu.sg.}
%\thanks{Yaqing Hou is with the College of Computer Science and Technology, Dalian University of Technology, China. Email: houyq@dlut.edu.cn.}
\thanks{Abhishek Gupta, Agency for Science, Technology and Research (A*STAR) \& Nanyang Technological University (NTU), SINGAPORE. Email: abhishek\_gupta@simtech.a-star.edu.sg.}
\thanks{Lei Zhou, NTU, SINGAPORE. Email: lei.zhou@ntu.edu.sg.}
\thanks{Yew-Soon Ong, A*STAR \& NTU, SINGAPORE. Email: asysong@ntu.edu.sg}
\thanks{Zefeng Chen, NTU, SINGAPORE. Email: zefeng.chen@ntu.edu.sg.}
\thanks{Yaqing Hou, Dalian University of Technology, CHINA. Email: houyq@dlut.edu.cn.}
\thanks{*\emph{Corresponding author}}
}


\markboth{Journal of \LaTeX\ Class Files,~Vol.~X, No.~X, XX~XX}%
{Shell \MakeLowercase{\textit{et al.}}: Bare Demo of IEEEtran.cls for IEEE Journals}



\maketitle
%

\noindent\begin{abstract}
Until recently, the potential to transfer evolved skills across distinct optimization problem instances (or tasks) was seldom explored in evolutionary computation. The concept of \emph{evolutionary multitasking} (EMT) fills this gap. It unlocks a population's implicit parallelism to jointly solve a set of tasks, hence creating avenues for skills transfer between them. Despite it being early days, the idea of EMT has begun to show promise in a range of real-world applications. In the backdrop of recent advances, the contribution of this paper is twofold. First, a review of several application-oriented explorations of EMT in the literature is presented; the works are assimilated into half a dozen broad categories according to their respective application domains. Each of these \emph{six} categories elaborates fundamental motivations to multitask, and contains a representative experimental study (referred from the literature). Second, a set of recipes is provided showing how problem formulations of general interest, those that cut across different disciplines, could be transformed in the new light of EMT. Our discussions emphasize the many practical use-cases of EMT, and is intended to spark future research towards crafting novel algorithms for real-world deployment.
\end{abstract}

% \fi
\IEEEpeerreviewmaketitle

\section{Introduction}
\noindent Optimization is at the heart of problem-solving. Many practical problems however possess non-convex, non-differentiable, or even black-box objectives and constraints that lie outside the scope of traditional mathematical methods. Evolutionary algorithms (EAs) provide a gradient-free path to solve such complex optimization tasks, with flexibility to cope with additional challenges such as expensive-to-evaluate objectives \cite{EOP}, dynamics \cite{DOP}, etc. EAs are population-based methods inspired by Darwinian principles of natural evolution, but, notably, fall short of simulating the phenomenon in its entirety \cite{bioEC}. Unlike the tendency of natural evolution to speciate or produce differently skilled sub-populations, the update mechanisms of \emph{in silico} EAs are usually crafted to evolve a set of solutions for only a single target task. This naturally limits the power of a population's \emph{implicit parallelism} \cite{MultiX}, often slowing down convergence rates as useful skills from other \emph{related} tasks are not readily accessible. The concept of \emph{evolutionary multitasking} (EMT) addresses this limitation by offering a new perspective on the potential of EAs.

The notion of generalizing beyond the ambit of just a single task is set to mould the future of search and optimization algorithms, especially since real-world problems seldom exist in isolation \cite{yao2021self, insights}. In scientific and engineering applications, for example, building on known knowledge and existing solutions can greatly reduce the time taken to explore and innovate new designs---which could otherwise take days, weeks, or even months to discover if probed in a tabula rasa manner \cite{min2020generalizing}. Yet, EAs continue to be crafted to work on problem instances independently, ignoring useful information gleaned from the solving of others. The notion of EMT fills this gap, launching the inter-task transfer and adaptive reuse of information across distinct, but possibly related, tasks. The transfer is achieved by unlocking a population's implicit parallelism in a new class of EAs equipped to jointly tackle multiple tasks.

EMT was put forward in \cite{MFO}, and has since attracted much interest among evolutionary computation (EC) researchers. A variety of algorithmic realizations have been proposed, including the single-population \emph{multifactorial EA} (MFEA) \cite{MFO}, multi-population algorithms \cite{multipop}, or other co-evolutionary algorithms \cite{coev}, aiming for efficient solving of multiple tasks by maximally utilizing mutual relationships through information transfer. To this end, research questions in terms of \emph{what, how, and when} to transfer arise in the unique context of EMT. In what follows, a brief overview of the ways in which today's EMT and transfer EAs address some of these questions is provided. Since an in-depth methodological analysis is not the focus of this paper, readers are also referred to \cite{survey2, survey1, wei2021review} for more comprehensive discussions on these topics.

Determining \emph{what} to transfer emphasises the type of information unit and its computational representation \cite{memetic}. Besides \emph{genetic transfers} of complete solution prototypes or subsets of solution strings (e.g., frequent schema) \cite{zheng2019self,zhou2020toward}, other knowledge representations have included probabilistic search distribution models \cite{memetic}, search direction vectors \cite{yin2019}, higher-order heuristics \cite{heuristic}, and surrogate models of expensive objective functions \cite{multiproblem}. Given the information type, \emph{how} to transfer becomes crucial when dealing with heterogeneous tasks (e.g., with differing search space dimensionality). Various solution representation learning strategies for mapping tasks to a common space have been proposed in this regard \cite{LMFEA,solre,SA1,SA2,chen2020learning}, with an abstract categorization of associated strategies presented in \cite{lim2021non}.

Given \emph{what} and \emph{how}, discerning situations \emph{when} to (or when not to) transfer is a natural follow-up to maximize utilization of inter-task relations while curbing harmful interactions. Increasing efforts have thus been made to craft adaptive EMT algorithms capable of online discovery of similarities even between black-box optimization tasks. The gleaned similarity has then been used to control on-the-fly the \emph{extent} of transfer between tasks \cite{MFEAII}, as opposed to earlier approaches that predefined and fixed this quantity \cite{MFO, EEMT}.

Ongoing works in EMT are deeply focused on addressing theoretical questions of the aforementioned kind, often assuming synthetic settings with algorithmic tests run on idealized benchmark functions. A mathematical proof of faster convergence has also been derived under convexity conditions \cite{bai2021multitask}. Given the methodological advances being made, the time is deemed ripe to also draw the attention of researchers and practitioners to the rich but nascent space of real-life applications of EMT. From the design of multiphysics products \cite{rios2021multi} to social network reconstruction \cite{shen2021evolutionary, wu2021evolutionary} and search-based software optimization \cite{Branch}, EMT promises significant performance gains in varied domains where multiple related problem instances routinely occur/recur. Thus, with the goal of strengthening the bridge between the theory and practice of EMT, this paper makes the following twofold contribution.
\begin{itemize}
\item An extensive literature review from the perspective of real-world applications of EMT is presented. Application-oriented explorations of multitasking are encapsulated in half-dozen broad categories together with representative experimental studies from past publications. Although by no means comprehensive, these six examples showcase the computational advantages that EMT could bring to areas such as the evolution of embodied intelligence, the path planning of unmanned vehicles, or last-mile logistics optimization, to name just a few.
\item Beyond the six application domains, the paper presents recipes by which certain problem formulations of applied interest---those that cut across disciplines---may be transformed in the new light of EMT. The formulations fall under the umbrella of \emph{multi-X} EC \cite{MultiX}, unveiling avenues by which a population's implicit parallelism, augmented by the capacity to multitask, could be further leveraged for practical problem-solving.
\end{itemize}
These discussions are intended to highlight the potential use-cases of EMT methods known today. In addition, it is hoped to spark future research toward innovative multitask algorithms crafted specifically for real-world deployment.

The rest of the paper is organized as follows. Section \ref{S2} introduces multitask optimization, formulates EMT from a probabilistic viewpoint, and presents a brief methodological overview. Section \ref{S3} sets out the half-dozen broad categories summarizing practical exemplars of EMT. Future applied research prospects of multitasking are then discussed in Section \ref{S4}. Section \ref{S5} concludes the paper.

\section{Background} \label{S2}
\noindent In this section, the preliminaries of multitask optimization are presented, a general probabilistic formulation for evolutionary multitasking is introduced, and representative algorithmic approaches from the literature are discussed---thus laying the foundation for the study of real-world applications of EMT.

\subsection{The Multitask Optimization Problem}
\noindent Multitask optimization (MTO) comprises multiple problem instances to be jointly solved. Without loss of generality, MTO consisting of \textit{K} tasks can be defined as\footnote{Only single-objective maximization is depicted for brevity. The concept of MTO readily extends to multiple multi-objective optimization tasks \cite{MOMFO}, or even a mixture of single- and multi-objective optimization tasks \cite{ma2021enhanced}. For minimization, fitness functions are simply multiplied by -1.}:
\begin{equation} \label{Eq:OP}
   \mathbf{x}^*_i = \mathop{\arg\max}_{\mathbf{x}_i \in \mathcal{X}_i}f_i(\mathbf{x}_i), \text{for} \;i = 1, 2, \dots, K,
\end{equation}
where $\mathbf{x}^*_i$, $\mathcal{X}_i$ and \textit{f$_i$} represent the optimal solution, search space, and objective/fitness function of the \textit{i}th task, respectively. Typically, optimization includes additional constraint functions, but these have been omitted in Eq. (\ref{Eq:OP}) for simplicity.

The motivation behind MTO is to facilitate the transfer of learned skills from one task to another, enhancing overall optimization performance. For such transfer to take place, a unified space $\mathcal{X}$ encoding candidate solutions for all $K$ tasks is usually defined (alternative approaches are discussed in Section \ref{Sec2C}). Let the encoding be achieved by an invertible mapping function $\psi_i$ for the \textit{i}th task, such that $\psi_i: \mathcal{X}_i \to \mathcal{X}$. Then, the decoding of solutions from the unified space back to a task-specific search space is given as $\psi_i^{-1}: \mathcal{X} \to \mathcal{X}_i$. Early works utilized the \emph{random-key} representation \cite{MFO} for solution mapping. More recently, solution representation learning strategies have been derived \cite{LMFEA, solre}, forming common highways by which building-blocks of knowledge derived from heterogeneous tasks (i.e., with differing search spaces) can be transmitted and recombined.
\subsection{A Probabilistic Formulation of EMT}
\noindent Defining an optimization task, with fitness function \textit{f$_i$} : $\mathcal{X}_i \to \mathbb{R}$, in terms of the \emph{expected fitness} under a probability distribution function gives \cite{wierstra2014natural}:
\begin{equation} \label{Eq:ST}
   \max \limits_{p_i(\mathbf{x}_i)} \int_{\mathcal{X}_i} f_i(\mathbf{x}_i)\cdot p_i(\mathbf{x}_i)\cdot d\mathbf{x}_i. \\
\end{equation}
In probabilistic model-based EAs, $p_i(\mathbf{x}_i)$ represents the underlying search distribution of a population of evolving solutions \cite{zhang2004convergence}. Note, if  $\mathbf{x}^*_i$ is the true optimum, then a Dirac delta function centred at $\mathbf{x}^*_i$ optimizes Eq. (\ref{Eq:ST}); i.e., $p^*_i(\mathbf{x}_i) = \delta(\mathbf{x}_i - \mathbf{x}^*_i)$ since $\int_{\mathcal{X}_i} f_i(\mathbf{x}_i)\delta(\mathbf{x}_i - \mathbf{x}^*_i) d\mathbf{x}_i = f_i(\mathbf{x}^*_i)$. As such, probabilistic reformation does not change the optimization outcome.

Consider MTO with \textit{K} tasks. These are encoded in a unified space $\mathcal{X}$ with a set of probabilistic models \{$p_1(\mathbf{x})$, $p_2(\mathbf{x})$, \dots, $p_K(\mathbf{x})$\} corresponding to task-specific (sub-)populations. One way to pose EMT is then by generalizing Eq. (\ref{Eq:ST}) over all $K$ tasks using probability mixture models as \cite{memetic}:
\begin{equation} \label{Eq:EMT}
\begin{split}
   \max \limits_{\{w_{ij}, p_j(\mathbf{x}), \forall i,j\}} & \sum_{i=1}^K \int_{\mathcal{X}} f_i(\psi_i^{-1}(\mathbf{x}))\cdot [\Sigma_{j=1}^K w_{ij}\cdot p_j(\mathbf{x})]\cdot d\mathbf{x}, \\
   s.t.~~ & \Sigma_{j=1}^K w_{ij}=1, \forall i, \\
   & w_{ij} \ge 0, \forall i,j,\\
\end{split} \end{equation}
where \textit{w}$_{ij}$'s are scalar coefficients indicating how individual models are assimilated into the mixture.

Note that Eq. (\ref{Eq:EMT}) would be exactly solved when all probabilistic models converge to the respective optimal Dirac delta functions $p^*_j(\mathbf{x}) = \delta(\mathbf{x}-\psi_j(\mathbf{x}^*_j))$ in $\mathcal{X}$, and $w_{ij} = 0$ for all $i \neq j$. Hence, the reformulation is in alignment with the definition of MTO in Eq. (\ref{Eq:OP}). \emph{More importantly however, by viewing the 'process' of multitasking through the lens of Eq. (\ref{Eq:EMT}), it is possible to adaptively control the extent of skills transfer between tasks by tuning the coefficients of the mixture models}. Precisely, if candidate solutions evolved for the \textit{j}th task---i.e., drawn from the model $p_j(x)$---are found to be performant for the \textit{i}th task as well, then the value of \textit{w}$_{ij}$ is increased to intensify the cross-sampling of solution prototypes. In contrast, if solutions transferred from a given \emph{source} do not excel in the recipient \emph{target} task, then the coefficient corresponding to that source-target pair is gradually neutralized. A complete algorithmic implementation of this general idea is described in \cite{memetic}.

\subsection{An Overview of EMT Methodologies}\label{Sec2C}
\noindent A plethora of EMT algorithms have been proposed lately. Some of these either directly or indirectly make use of the formulation in Eq. (\ref{Eq:EMT}). Nevertheless, as noted in \cite{survey2}, most algorithms typify one of the two methodological classes stated below. An extensive analysis of these methods is not included herein as excellent reviews are available elsewhere \cite{survey1, wei2021review}; only a handful of representative approaches are discussed.

(1) \emph{EMT with implicit transfer:} In these methods, the exchange of information between tasks occurs through evolutionary crossover operators acting on candidate solutions of a \emph{single population} \cite{twostage,MFPSODE,GrMFEA}. The population is encoded in a unified space $\mathcal{X}$ using task-specific invertible mapping functions $\psi_i$. Implicit genetic transfers materialize as solutions evolved for different tasks crossover in $\mathcal{X}$, hence exchanging learnt skills coded in their genetic material. Over the years, a multitude of evolutionary crossover operators have been developed, each with their own biases. The success of implicit transfers between any task pair thus depends on whether the chosen crossover operator is able to reveal and exploit relationships between their objective function landscapes. For example, in \cite{landscape}, an \emph{offline} measure of inter-task correlation was defined and evaluated for parent-centric crossovers synergized (strictly) with gradient-based local search updates. In \cite{MFEAII}, an \emph{online} measure was derived by means of a latent probability mixture model, akin to Eq. (\ref{Eq:EMT});  the mixture was shown to result from the use of parent-centric operators in the single-population MFEA. (Adapting the extent of transfer based on the coefficients of the mixture model then led to the MFEA-II algorithm.) Greater flexibility in operator selection can however be achieved through self-adaptation strategies as proposed in \cite{zhou2020toward}, where data generated during evolutionary search is used for online identification of effective crossover operators for transfer.

(2) \emph{EMT with explicit transfer:} Here, information exchanges take place between \emph{multiple populations}. Each population corresponds to a task in MTO and evolves in problem-specific search space $\mathcal{X}_i, \forall i$. The populations mostly evolve independently, between periodic stages of knowledge transfer. An explicit transfer mechanism is triggered whenever a predefined condition, e.g., transfer interval, is met \cite{EEMT}. For homogeneous cases where $\mathcal{X}_1=\mathcal{X}_2=\dots =\mathcal{X}_K$, island-model EAs for multitasking have been proposed \cite{hashimoto2018analysis}, with added functionality to control the frequency and quantity of solution cross-sampling \cite{EBS}. Under heterogeneous search spaces, mapping functions $\psi_{ij} : \mathcal{X}_i \to \mathcal{X}_j$, for all $i \neq j$, must be defined to reconcile $i$th and $j$th populations. To this end, while most existing EMT methods have made use of linear mapping functions \cite{EEMT, lin2019IL}, the applicability of fast yet expressive nonlinear maps, as proposed for \emph{sequential} transfers in \cite{lim2021non, zhou2021learnable}, are deemed worthy of future exploration.

Both methodological classes have their merits. In the spirit of the no free lunch theorem \cite{wolpert1997no}, algorithmic design preferences must therefore be guided by the attributes of the application at hand. Note that implicit genetic transfers naturally emerge from standard evolutionary operations in unified space, without having to craft ad hoc transfer mechanisms; hence, implementation is relatively simple and scales well for large $K$. However, composing a unified space and search operators for heterogeneous tasks becomes highly non-trivial (operators that work well for one task may not be effective for another). In contrast, the multi-population approach of explicit transfer suppresses the need for unification, allowing each task to hold specialized search operators. But additional complexity may be introduced in having to define $\mathcal{O}(K^2)$ inter-task solution mappings ($\psi_{ij}$'s) under heterogeneous search spaces \cite{wei2021review}.

\section{A Review of EMT in Action in Real-World Problems} \label{S3}
\noindent The aim of this section is to draw the attention of both researchers and practitioners to the many practical use-cases of EMT. Prior literature exploring real-world applications is encapsulated in six broad categories, together with representative case studies and published results that showcase its benefits.

\subsection{Category 1: EMT in Data Science Pipelines}
\noindent Many aspects of data science and machine learning (ML) pipelines benefit from the salient features of EAs for optimization. Problems such as feature selection \cite{xue2015survey}, hyper-parameter tuning \cite{huang2019surrogate}, neural architecture search \cite{liu2021survey}, etc., involve non-differentiable, multimodal objective functions and discrete search spaces that call for gradient-free optimization. Population-based EAs have even been considered as worthy rivals to, or in synergy with, stochastic gradient descent for learning with differentiable loss functions \cite{morse2016simple, cui2018evolutionary}. Despite the advances, there remain challenges in the efficient scaling of EAs to scenarios characterized by big data (e.g., containing a large number of individual data points), large-scale (high-dimensional) feature/parameter spaces, or involving building sets of multiple learning algorithms (e.g., ensemble learning). EMT provides different pathways to sustain the computational tractability of EAs in such data science settings.

\emph{EMT with auxiliary task generation:} Approaches to augment the training of ML models by turning the problem into MTO---with artificially generated \emph{auxiliary tasks}---were introduced in \cite{zhangtraining}. In neural networks for instance, tasks could be defined by different loss functions or network topologies, with the transfer of model parameters between them leading to better training \cite{chandra2018evolutionary}. More generally, for scaling-up the evolutionary configuration of arbitrary ML subsystems, the idea of constructing auxiliary \emph{small data} tasks from an otherwise large dataset was proposed  in \cite{Zhang2021Minions, wang2022evolutionary}. The auxiliary tasks can then be combined with the main task in EMT, accelerating search by using small data to quickly optimize for the large dataset; speedups of over 40\% were achieved in some cases of wrapper-based feature selection via an EMT algorithm with explicit transfer \cite{Zhang2021Minions}. In another application for feature selection, the tendency of stagnation of EAs in high-dimensional feature spaces was lessened by initiating information transfers between artificially generated low-dimensional tasks \cite{chen2020evolutionary, chen2021evolutionary}.

\begin{figure}
  \centering
  \includegraphics[width=0.42\textwidth]{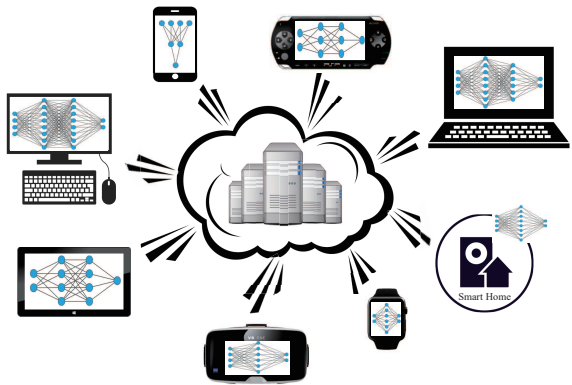}
  \caption{Cloud computing platforms house black-box optimization services where users can simply upload their raw data to have optimized predictive models delivered \cite{golovin2017google}. In this setting, EMT could harness knowledge transfers across non-identical but related tasks (e.g., with different training data and/or device requirements) to enable efficient model configuration.}
  \label{Fig:GCloud}
\end{figure}

\emph{EMT on sets of learning algorithms:} Given a training dataset, an ensemble (or set) of classification models could be learnt by simple repetition of classifier evolution. However, this would multiply computational cost. As an alternative, the study in \cite{wen2016learning} proposed a variant of \emph{multifactorial genetic programming} (MFGP) for simultaneous evolution of an ensemble of decision trees. MFGP enabled a set of classifiers to be generated in a single run, with the transfer and reuse of common subtrees providing substantial cost savings in comparison to repeated (independent) runs of genetic programming. Moving upstream in the data science pipeline, \cite{zhang2018evolutionary} formulated the task of finding optimal feature subspaces for each base learner in an ensemble as an MTO problem. An EMT feature selection algorithm was then proposed to solve this problem, yielding feature subspaces that often outperformed those obtained by seeking the optimal for each base learner independently. A similar idea but targeting the specific case of hyperspectral image classifiers was presented in \cite{shi2020evolutionary}.

Beyond the training of ML models, the literature evinces applications of EMT for image processing as well. For sparse unmixing of hyperspectral images, the approaches in \cite{li2018evolutionary, zhao2020endmember} suggest to first partition an image into a \emph{set} of homogeneous regions. Each member of the set is then incorporated as a constitutive sparse regression task in EMT, allowing implicit genetic transfers to exploit similar sparsity patterns. Results revealed faster convergence to near-optimal solutions via EMT, as opposed to processing pixels or groups of pixels independently. In \cite{li2021multi}, a multi-fidelity MTO procedure was incorporated into the hyperspectral image processing framework. A surrogate model was used to estimate the gap between low- and high-fidelity evaluations to achieve further improvements in accuracy and algorithmic efficiency.

\emph{EMT across non-identical datasets:} It is envisaged that future cloud-based black-box optimization services shall open up diverse applications of EMT for automated configuration of learning algorithms. Comparable services are already on the horizon, making it possible for users to upload their raw data to the cloud and have high-quality predictive models delivered without the need for extensive human inputs \cite{golovin2017google}. Different user groups may possess \emph{non-identical} data, and, as depicted in Fig. \ref{Fig:GCloud}, may even pose different device requirements constraining the transition of trained models from the cloud to the edge. In such settings, EMT could effectively function as an expert ML practitioner, exploiting knowledge transfers across non-identical but related domains to speed up model configuration. Early works showing the plausibility of this idea---using a distinct class of multitask Bayesian optimization algorithms---were presented in \cite{swersky2013multi, adams2018systems}.

Recently, an intriguing application of EMT feature selection to understand the employability of university graduates has been explored \cite{jayaratnaunderstanding}. Students studying different disciplines (business, engineering, etc.) formed multiple non-identical cohorts, with the data for each cohort forming a feature selection task in MTO. Then, by allowing common features/attributes to be shared through multitasking, efficient identification of determinants that most influence graduate employment outcomes was achieved. In  \cite{bi2020learning}, a multitask genetic programming algorithm for feature learning from images was proposed. For a given pair of related but non-identical datasets, the approach jointly evolves common trees together with task-specific trees that extract and share higher-order features for image classification. The effectiveness of the approach was experimentally verified for the case of simultaneously solving two tasks, showing similar or better generalization performance than single-task genetic programming.

\newsavebox{\tablebox}
\begin{lrbox}{\tablebox}%
\renewcommand\arraystretch{1.5}
\begin{tabular}{c l c c }
  \hline
  \hline
  \multicolumn{2}{c}{} & CO$_2$   & DRP \\
    \hline
    \multirow{5}{*}{MFGP} & paired problem & $RMSE$  & $RMSE$ \\
    \cline{2-4}
          & CO$_2$   & N/A   & 0.494 \\
    \cline{2-4}
          & S\_CO$_2$ & \textbf{4.828} & N/A \\
    \cline{2-4}
          & DRP   & 5.495 & N/A \\
    \cline{2-4}
          & S\_DRP & N/A   & \textbf{0.478} \\
    \hline
    \multicolumn{2}{c}{SL-GEP} & 5.504 & 0.534\\
  \hline
  \hline
\end{tabular}
\end{lrbox}
\begin{table}
\centering
\small \caption{RMSE values achieved by MFGP and single-task SL-GEP for the symbolic regression of time series data. Best values are marked in bold. The results are obtained from \cite{SRP}.  }\label{Table-SRP} \scalebox{0.95}{\usebox{\tablebox}}
\end{table}

\begin{figure*}
  \centering
  \includegraphics[width=0.77\textwidth]{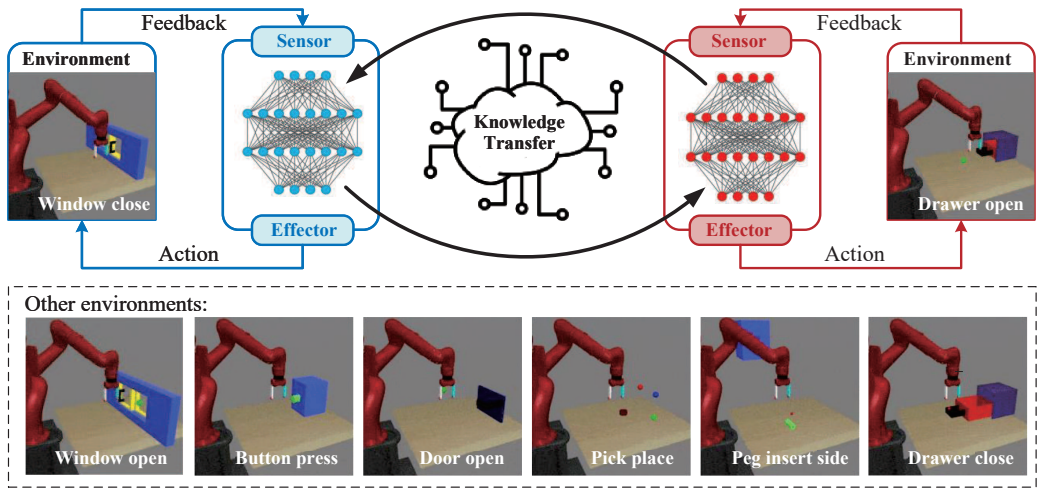}
  \caption{The \emph{window close} and \emph{drawer open} tasks share similar approaching and pulling movements. Hence, training a robot to perform such tasks simultaneously via EMT allows mutually beneficial knowledge transfers to occur. The lower figure panel visualizes the same robot situated in other Meta-World environments that were included in the experimental study in \cite{martinez2021adaptive}.}
  \label{Fig:RL}
\end{figure*}

\begin{spacing}{1.5}
\noindent \emph{$\bullet$ Case study in symbolic regression modeling \cite{SRP}}
\end{spacing}

\noindent Many works in the literature have explored multitasking in connection with genetic programming \cite{scott2017multitask, scott2019automating}. Here, a real-world study of MFGP comprising two symbolic regression tasks with distinct time series data is considered \cite{SRP}.

The first problem instance contains 260 data points representing monthly average atmospheric CO$_2$ concentrations collected at Alert, Northwest Territories, Canada from January 1986 to August 2007. The second problem instance contains 240 data points representing monthly U.S. No 2 Diesel Retail Prices (DRP) from September 1997 to August 2017. Two simplified tasks with reduced time series datasets were also generated by subsampling of the original data. These were labelled as S\_CO$_2$ and S\_DRP, respectively. The MFGP was thus applied to solve three pairs of tasks, i.e., \{CO$_2$, S\_CO$_2$\}, \{CO$_2$, DRP\} and \{DRP, S\_DRP\}, each with the goal of deriving a symbolic (closed-form mathematical) equation mapping elapsed time to the output prediction. Equations were evolved by minimizing their root mean square error ($RMSE$) \cite{SRP}.

Table \ref{Table-SRP} summarizes the $RMSE$ values obtained by MFGP and its single-task counterpart SL-GEP \cite{zhong2015self}. Superior results are highlighted in bold.
As can be seen, MFGP outperformed SL-GEP in all experimental settings. Particularly, the best results of CO$_2$ and DRP were achieved when paired with their corresponding simplified problem variants. This is intuitively agreeable as the simplified tasks (generated by subsampling) are expected to be similar to the original problem instances, hence engendering fruitful transfers of genetic building-blocks that speed up convergence and improve performance.

\subsection{Category 2: EMT in Evolving Embodied Intelligence}

\noindent Evolutionary robotics takes a biologically inspired view of the design of autonomous machines \cite{bongard2013evolutionary}. In particular, EAs are used to adapt robots/agents to their environment by optimizing the parameters and architecture of their control policy (i.e., the function transforming their sensor signals to motor commands) while accounting for, or even jointly evolving, the morphology of the agent itself. It is the design of intelligent behaviour through this interplay between an agent and its environment, mediated by the physical constraints of the agent’s body, sensory and motor system, and brain, that is regarded as \emph{embodied intelligence} \cite{cangelosi2015embodied}. Put differently, while mainstream robotics seeks to generate better behaviour for a given agent, embodied intelligence enables agents to adapt to diverse forms, shapes and environments, hence setting the stage for the efficacy of EMT with implicit or explicit genetic transfer to be naturally realized \cite{moshaiov2014family}.

Imagine evolving embodied intelligence by means of different tasks parameterized by an agent's morphological and environmental descriptors. In \cite{kinematic-arm}, a multitasking analogue of an archive-based exploratory search algorithm \cite{mouret2015illuminating} was used to train a 6-legged robot to walk forward as fast as possible under different morphologies derived by changing the lengths of its legs. Each set of lengths thus defined a specific task. The experiments evolved walking gait controllers for 2000 random morphologies (or tasks) at once, under the intuition that a particular controller might transfer as a good starting point for several morphologies. The results successfully substantiated this intuition, showing that a multitask optimization algorithm was indeed able to significantly outperform a strong single-task baseline.

Similarly, in \cite{kinematic-arm} and \cite{wang2021solving}, a set of planar robotic arm articulation tasks with variable morphology were formulated by parameterizing the arm by the length of its links. The objective of each task was then to find the angles of rotation of each joint minimizing the distance between the tip of the arm and a predefined target. The experiments in \cite{wang2021solving} confirmed that an anomaly detection model-based adaptive EMT algorithm, with explicit transfer strategy, could achieve faster convergence and better objective function values (averaged across all tasks) compared to the baseline single-task EA.

While the two previous examples considered robot morphological variations, \cite{martinez2021adaptive} applied EMT (in particular, an adaptive version of the MFEA) for simulation-based deep learning of control policies of a robot arm situated in different \emph{Meta-World} environments \cite{yu2020metaworld}. As shown in Fig. \ref{Fig:RL}, the various tasks in MTO involved deep neuroevolution of policy parameters of a robot arm interacting with different objects, with different shapes, joints, and connectivity. In the experiments, up to 50 tasks were evolved at the same time, with crossover-based exchange of skills between synergistic tasks leading to higher success rates as well as lower computational cost compared to a single-task soft actor critic algorithm \cite{martinez2021adaptive}.

\begin{spacing}{1.5}
\noindent \emph{$\bullet$ Case study in neuroevolution of robot controllers \cite{MFEAII}}
\end{spacing}

\begin{lrbox}{\tablebox}%
\renewcommand\arraystretch{1.5}
\begin{tabular}{c c|c c c c c}
  \hline
  \hline
  \multirow{2}{*}{Task} & \multirow{2}{*}{$l_s$} & \multirow{2}{*}{CEA} & \multicolumn{4}{c}{MFEA-II} \\
  \cline{4-7}
  &&& \{$T_1, T_2$\}& \{$T_1, T_3$\}& \{$T_2, T_3$\}& \{$T_1, T_2, T_3$\}\\
  \hline
  $T_1$ & 0.60m & 27\% & 30\% & 30\% & - & \textbf{47}\%\\
  \hline
  $T_2$ & 0.65m & 0\% & 27\% & - & 27\% & \textbf{37}\%\\
  \hline
  $T_3$ & 0.70m & 0\% & - & 7\% & \textbf{27}\% & 17\%\\
  \hline
  \hline
\end{tabular}
\end{lrbox}
\begin{table}
\centering
\small \caption{Comparison of success rates (in \%) achieved by MFEA-II and a single-task canonical EA (CEA) on different double pole balancing problem instances. Results are obtained from \cite{MFEAII}.}\label{Table-DBP1} \scalebox{0.8}{\usebox{\tablebox}}
\end{table}

\noindent Here, a case study on the classical \emph{double pole balancing} problem under morphological variations is considered. The basic problem setup consists of two inverted poles of different lengths hinged on a moving cart. The objective is for a neural network controller to output a force that acts on the moving cart such that both poles are balanced (i.e., remain within an angle of $\pm 36^{\circ}$ from the vertical for a specified duration of simulated time), while also ensuring that the cart does not go out of bounds of a 4.8 m horizontal track. Neuroevolution of network parameters continues until the poles are successfully balanced, or the available computational budget is exhausted.  The success rates of EAs over multiple randomly initialized runs are recorded for comparison. The input to the neural network is the state of the system which is fully defined by six variables: the position and velocity of the cart on the track, the angle of each pole from the vertical, and the angular velocity of each pole. The Runge-Kutta fourth-order method is used to simulate the entire system.

Multiple morphologies in MTO were constructed by varying the difference in the lengths of the two poles. In particular, the length of the long pole was fixed at 1 m, while the length $l_s$ of the shorter pole was set as either 0.60 m ($T_1$), 0.65 m ($T_2$), or 0.70 m ($T_3$). Four resulting MTO settings are denoted as \{$T_1, T_2$\}, \{$T_1, T_3$\}, \{$T_2, T_3$\}, and \{$T_1, T_2, T_3$\}. The architecture of the neural network controller (two-layer with ten hidden neurons) was kept the same for all tasks, thus providing an inherently unified parameter space for transfer. It is well-known that the double pole system becomes increasingly difficult to control as the length of the shorter pole approaches that of the long pole. However, by simultaneously tackling multiple tasks with different levels of difficulty, the controllers evolved for simpler tasks could transfer to help solve more challenging problem instances efficiently.

This intuition was borne out by the experimental studies in \cite{MFEAII}, results of which are also depicted in Table \ref{Table-DBP1}. A single-task canonical EA (CEA) could only achieve a success rate of 27\% on task $T_1$ while failing on the more challenging instances $T_2$ and $T_3$. In contrast, the MFEA-II algorithm, equipped with exactly the same operators as CEA, achieved better performance across \emph{all} tasks by virtue of unlocking inter-task skills transfer. Not only did the success rate of $T_1$ reach 47\% (indicating that useful information could even transfer from challenging to simpler tasks), but that of $T_2$ and $T_3$ also reached a maximum of 37\% and 27\%, respectively.

\subsection{Category 3: EMT in Unmanned Systems Planning}

\begin{figure}
  \centering
  \includegraphics[width=0.37\textwidth]{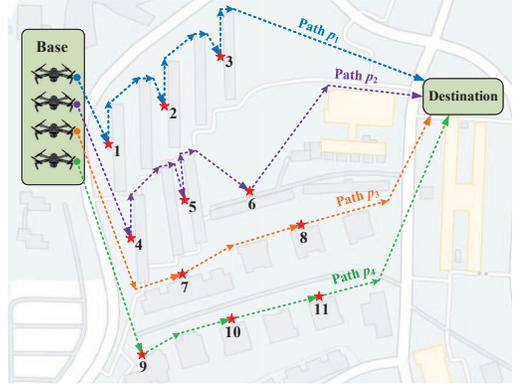}
  \caption{An illustration of multi-agent path planning. Red stars denote waypoints between the base station and the destination that must be visited by set of UAVs. The flight paths of different UAVs share similar, and hence transferrable, segments (such as segments 1-to-2 in path $p_1$ and 4-to-5 in path $p_2$, or segments 7-to-8 in path $p_3$ and 9-to-11 in path $p_4$) due to their similar surroundings (e.g., buildings).}
  \label{Fig:Robot}
\end{figure}

\noindent Evolutionary approaches are being used to optimize individual behaviours in robot swarms and unmanned vehicle systems. Consider unmanned aerial vehicles (UAVs) as an example. As their usage increases, UAV traffic management systems would be needed to maximize operational efficiency and safety \cite{rubio2018data}, avoiding catastrophes such as collisions, loss of control, etc. In such settings, each UAV may be viewed as an individual agent that perceives its surroundings to solve its corresponding task (e.g., path planning). The communication of acquired perceptual and planning information to other UAVs in related environments could then lead to better and faster decisions collectively. An illustration is depicted in Fig. \ref{Fig:Robot} where flight paths of different UAVs share similar straight or bent segments; these priors can be transferred and reused (as common solution building-blocks) to support real-time multi-UAV optimization. Explicit EMT offers a means to this end.

An early demonstration of this idea was presented in \cite{cogmt}, where two different multi-UAV missions were optimized jointly via the MFEA. The missions were optically distinct. While the first involved a pair of UAVs flying through two narrow openings in a barrier, the second involved four UAVs flying around a geofence of circular planform. The flight paths in both missions however possessed a hidden commonality. In all cases, the optimal magnitude of deviation from the straight line joining the start and end points of any UAV's path was the same. The MFEA successfully exploited this commonality to quickly evolve efficient flight paths.

A similar application was carried out in \cite{robot-path1} for the path planning of mobile agents operating in either the same or different workspaces. It was confirmed that EMT could indeed lead to the efficient discovery of workspace navigation trajectories with effective obstacle avoidance. In \cite{robot-path3}, a multi-objective robot path planning problem was considered to find solutions that optimally balance travel time and safety against uncertain path dangers. Given three topographic maps with distinct terrains, but bearing similarity in the distribution of obstacles, an EMT algorithm transferring evolved path information was shown to converge to sets of shorter yet safer paths quicker than its single-task counterpart.

\begin{figure}
  \centering
  \includegraphics[width=0.38\textwidth]{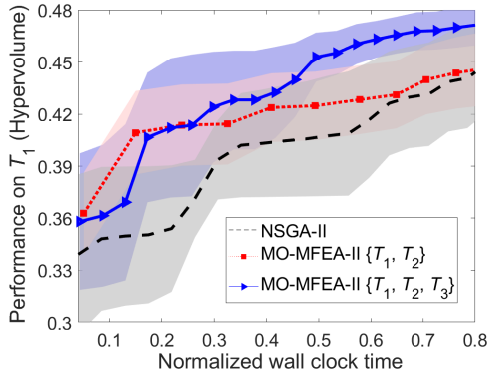}
  \caption{Convergence trends of NSGA-II and MO-MFEA-II on multi-UAV path planning. MO-MFEA-II incorporates lower-fidelity auxiliary tasks to help optimize the high-fidelity target $T_1$. Plots are obtained from \cite{robot-path2}. The shaded area spans $1/2$ standard deviation on either side of the mean performance.}
  \label{Fig:UAV}
\end{figure}

\begin{spacing}{1.5}
\noindent \emph{$\bullet$ Case study in safe multi-UAV path planning \cite{robot-path2}}
\end{spacing}

\noindent As a real-world example, a case study on the multi-objective path planning of five UAVs deployed in a $10\times7$ $km^{2}$ region in the southwest of Singapore is presented. The problem is characterized by uncertainty, stemming from the sparsity of data available to model key environmental factors that translate into operational hazards. The objective is thus to minimize travel distance while also minimizing the probability of unsafe events (which could be caused by flying through bad weather, or by loss of control due to poor communication signal strength). The latter objective is quantified based on a \emph{path-integral} risk metric derived in \cite{rubio2018data}. The resultant bi-objective optimization problem is further supplemented with constraint functions to ensure safe distance between UAVs, concurrence with altitude boundaries, and prevention of geofence breaches; refer to \cite{robot-path2} for a detailed description.

The ultimate goal of such a path planning system is to enable real-time decision support. However, the path-integral risk metric is computed via a numerical quadrature scheme that becomes computationally expensive for accurate risk estimation (i.e., when using a high-resolution 1D mesh). Hence, an MTO formulation was proposed in \cite{robot-path2} where cheaper low- and medium-fidelity auxiliary tasks were generated (by means of lower-resolution meshes) and combined with the main high-fidelity task at hand. The high-, medium-, and low-fidelity tasks are denoted as $T_1$, $T_2$ and $T_3$, respectively.

Fig. \ref{Fig:UAV} compares the optimization performance obtained by a single-task multi-objective EA \cite{deb2002fast} (solving just the high-fidelity task) and a multi-objective version of MFEA-II (MO-MFEA-II) \cite{robot-path2} solving \{$T_1$, $T_2$\} or \{$T_1$, $T_2$, $T_3$\}. The \emph{hypervolume} metric \cite{while2006faster} is used to quantify convergence trends in multidimensional objective space. As seen in the figure, both MO-MFEA-II settings led to better hypervolume scores faster than the  conventional single-task approach. The speedup is greater when given two auxiliary tasks (i.e., in the case of MTO with \{$T_1$, $T_2$, $T_3$\}), demonstrating the advantage of transferring good solutions generated by lower-fidelity tasks to quickly optimize the target problem instance.

\subsection{Category 4: EMT in Complex Design}

\begin{figure*}
  \centering
  \includegraphics[width=0.8\textwidth]{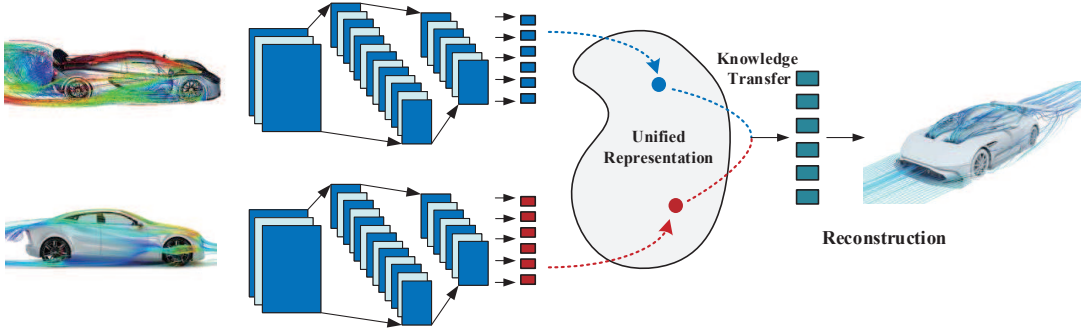}
  \caption{In many applications of EMT for engineering design, the lack of clear semantic overlap between design parameters could lead to difficulties in the construction of the unified search space $\mathcal{X}$. One example is in the definition of the unified space of diverse car shapes/geometries for aerodynamic design, which was addressed in \cite{rios2021multi} using a 3D point cloud autoencoder. Once trained, inter-task knowledge transfers take place in the latent space of the autoencoder.}
  \label{Fig:AERO}
\end{figure*}

\noindent The evaluation of solutions in scientific and engineering design domains often involves time-consuming computer simulation or complex laboratory procedures to be carried out (such as synthesizing candidate protein structures for protein optimization). The need for active solution sampling and evaluation to solve such tasks \emph{from scratch} can thus become prohibitively expensive. MTO provides an efficient alternative that has begun to attract widespread attention; examples of practical application have included finite element simulation-based system-in-package design \cite{dai2020sipd}, finite difference simulation-based optimization of well locations in reservoir models \cite{affine}, parameter identification of photovoltaic models \cite{photovoltaic},  optimization of active and reactive electric power dispatch in smart grids \cite{liu2020multitasking}, design of a coupled-tank water level fuzzy control system \cite{water-control}, to name a few. The hallmark of EMT in such applications lies in seeding transferred information into the search, hence building on solutions of related tasks to enable rapid design optimization. This attribute promises to particularly enhance the \emph{concpetualization phase} of design exercises, where multiple concepts with latent synergies are conceived and assessed at the same time \cite{cogmt, avigad2009interactive}.

Take car design as an exemplar. In \cite{car1, car2}, multifactorial algorithms were applied to simultaneously optimize the design parameters of three different types of Mazda cars---a sport utility vehicle, a large-vehicle, and a small-vehicle---of different sizes and body shapes, but with the same number of parts. (The three problem instances were first proposed in \cite{kohira2018proposal}, where the structural simulation software LS-DYNA\footnote{https://www.lstc.com/products/ls-dyna} was used to evaluate collision safety and build approximate response surface models.) Each car has 74 design parameters representing the thickness of the structural parts for minimizing weight while satisfying crashworthiness constraints. The experimental results in \cite{car1} showed that EMT was able to achieve better performance than the conventional (single-task) approach to optimizing the car designs. In another study, multitask shape optimization of three types of cars---a pick-up truck, a sedan, and a hatchback---was undertaken to minimize aerodynamic drag (evaluated using OpenFOAM\footnote{https://www.openfoam.com/} simulations) \cite{rios2021multi}. The uniqueness of the study lies in using a 3D point cloud autoencoder to derive a common design representation space (fulfilling the role of $\mathcal{X}$ in Eq. (\ref{Eq:EMT})) that unifies different car shapes; a graphical summary of this idea is depicted in Fig. \ref{Fig:AERO}. The transfer of solution building-blocks through the learnt latent space not only opened up the possibility of ``out of the box" shape generation, but also yielded up to 38.95\% reduction in drag force compared to a single-task baseline given the same computational budget \cite{rios2021multi}.

Not limited to structural and geometric design, EMT has also been successfully applied to \emph{process design} optimization problems. In an industrial research \cite{annealing-production}, an \emph{adaptive multi-objective, multifactorial differential evolution} (AdaMOMFDE) algorithm was proposed for optimizing continuous annealing production processes under different environmental conditions. A set of environmental parameters defined a certain steel strip production task, with multiple parameter sets forming multiple problem instances in MTO. Each task possessed three objectives, that of achieving prescribed strip hardness specifications, minimization of energy consumption, and maximization of production capacity.  Experiments simultaneously solving up to eight tasks were carried out in \cite{annealing-production}. The results demonstrated that the AdaMOMFDE algorithm could significantly outperform the single-task NSGA-II (as quantified by convergence trends of the inverted generational distance metric), hence meeting design specifications while potentially boosting productivity in the iron and steel industry.

In addition to the focused application areas above, MTO lends a general framework for handling expensive design optimizations by jointly incorporating tasks of multiple levels of fidelity. The real-world case study in the previous subsection was a case in point, albeit belonging to a different category. Other related studies have also appeared in the literature \cite{operational-indices}, a more extended discussion on which shall be presented in Section \ref{S4}-B of this paper.

\begin{spacing}{1.5}
\noindent \emph{$\bullet$ Case study in simulation-based process design \cite{MOMFO}}
\end{spacing}

\noindent This study showcases an example where EMT was applied to jointly optimize two types of liquid composite moulding (LCM) processes for producing the same lightweight  composite part \cite{MOMFO}. The part under consideration was a glass-fibre-reinforced epoxy composite disk, while the two LCM processes were resin transfer moulding (RTM) and injection/compression LCM (I/C-LCM). The process details are not reproduced herein for the sake of brevity; interested readers are referred to \cite{MOMFO}. The key characteristic of these two processes is that they possess \emph{partially overlapping} design spaces. Specifically, there exist three design parameters---the pressure and temperature of the epoxy resin when injected into the mould, and the temperature of the mould itself---that have similar physical effect on both LCM processes, hence leading to the scope of exploitable inter-task synergies.

The RTM and I/C-LCM optimization problem instances were formulated as bi-objective minimization tasks. The first objective was to minimize mould filling time (which in turn increases process throughput), while the second was to minimize peak internal fluid and fibre compaction force (which in turn reduces setup and running cost of peripheral equipment). For a set of candidate design parameters, the objective function values for either task were evaluated using a dedicated finite element numerical simulation engine.

\begin{figure}
  \centering
  \subfigure[]{
  \label{RTMcvg}
  \includegraphics[width=0.23\textwidth]{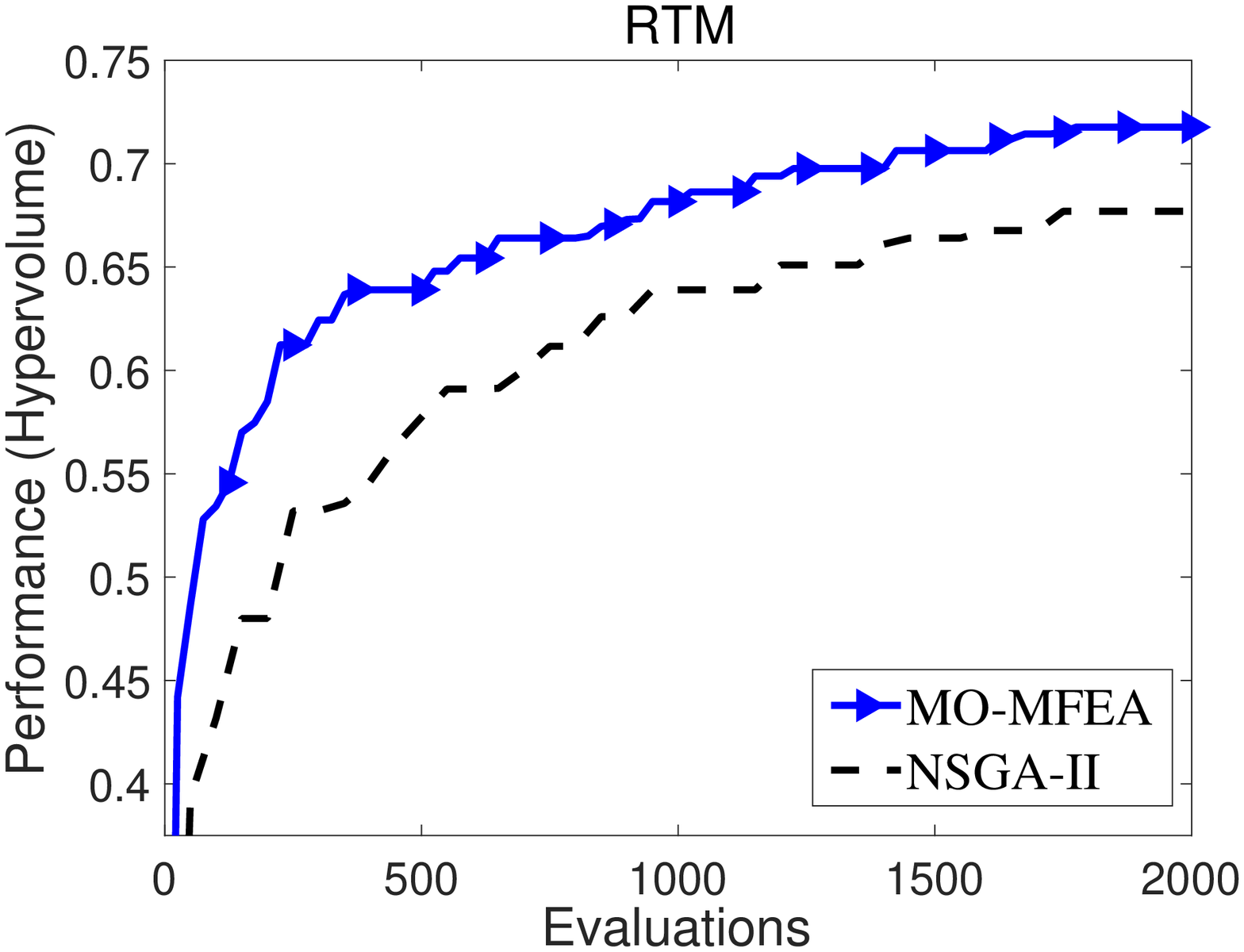}}
  \subfigure[]{
  \label{IC-LCMcvg}
  \includegraphics[width=0.23\textwidth]{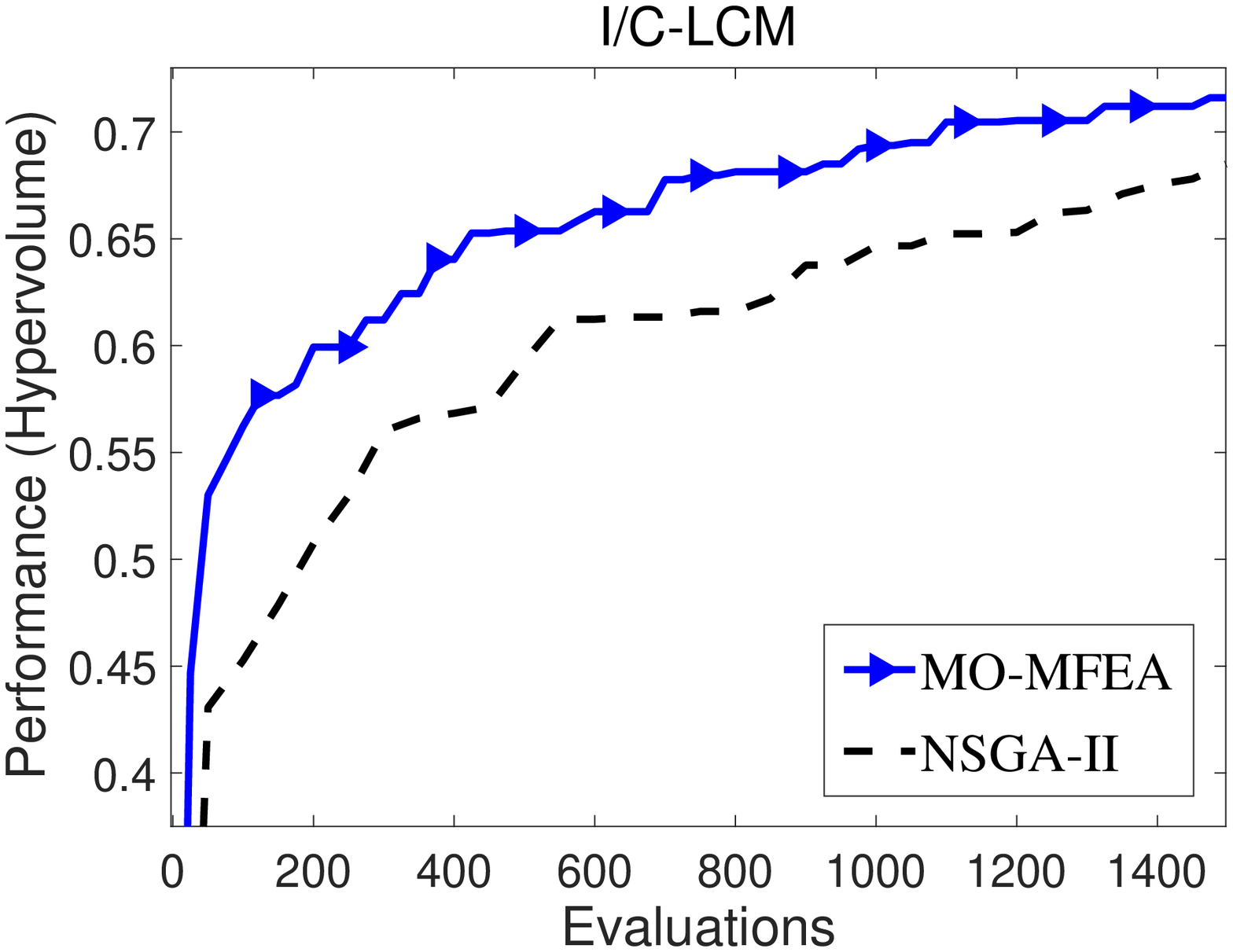}}
  \caption{(a) Hypervolume convergence trends of MO-MFEA and NSGA-II on the RTM process optimization task; (b) hypervolume convergence trends of MO-MFEA and NSGA-II on the I/C-LCM process optimization task. These plots have been obtained from the real-world study in \cite{MOMFO}.}
  \label{Fig:LCMcvg}
\end{figure}

\begin{figure*}
  \centering
  \includegraphics[width=0.82\textwidth]{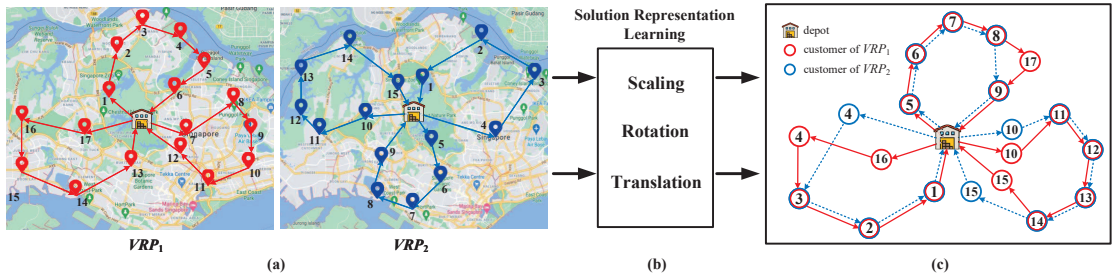}
  \caption{(a) $VRP_1$ and $VRP_2$ possess seemingly dissimilar node distribution and labels; (b) solution representation learning is undertaken to isometrically transform the node distribution of $VRP_2$ to match $VRP_1$; (c) the similarity of the two VRPs is unveiled after the transformation  \cite{solre}.}
  \label{Fig:CVRP}
\end{figure*}

The outputs of the multitasking MO-MFEA and the single-task NSGA-II are compared in Fig. \ref{Fig:LCMcvg} in terms of the normalized hypervolume metric. The convergence trends achieved by MO-MFEA on both tasks were found to surpass those achieved by NSGA-II. Taking RTM as an example (see left panel of Fig. \ref{Fig:LCMcvg}), the MO-MFEA took only about 1000 evaluations to reach the same hypervolume score reached by NSGA-II at the end of 2000 evaluations. This represents a $\sim$50\% saving in cost, which for expensive simulation-based optimization problems (ubiquitous in scientific and engineering applications) translates to substantial reduction in design time and the wastage of valuable physical resources.

\subsection{Category 5: EMT in Manufacturing, Operations Research}

\noindent The grand vision of smart manufacturing involves integration of three levels of manufacturing systems, namely, the shop floor, enterprise, and supply chain, into automated and flexible networks that allow for seamless data collection (via distributed sensors), data exchange, analysis, and decision-making \cite{tay2021model}. These may be supported by a \emph{nerve center} or \emph{manufacturing control tower}, where real-time data is collected across all system levels to offer centralized processing capacity and end-to-end visibility. It is in enabling effective functioning of such control towers that we foresee EMT to thrive, leveraging the scope of seamless data exchanges to deliver fast and optimal (or near-optimal) operational decisions \cite{jiang2016complex}.

Targeting energy efficient data collection and transmission to the base location (e.g., the nerve center), \cite{tam2021multifactorial} demonstrated the utility of EMT for optimizing the topology of wireless sensor networks. The optimization of both single-hop and multi-hop network types were combined in MTO to help with consideration of both deployment options. It was shown using a variant of the MFEA with random-key encoding that the exchange of useful information derived from solving both tasks could in fact lead to better overall results than the baseline single-task method. In \cite{huong2021multi}, the follow-on problem of charging the wireless sensors was also undertaken using a multitask approach. Multiple mobile chargers were simultaneously considered, with the charging schedule for each forming a task in MTO.

Returning to manufacturing operations, there exists a sizeable amount of research on applying EMT algorithms to NP-hard problems at the shop floor (e.g., for job shop scheduling \cite{DJSP, zhang2021surrogate}) or at the logistics and supply chain levels (e.g., for vehicle routing applications \cite{shang2022solving, osaba2020transferability} and its extension to pollution-routing \cite{PVRP}). For last-mile logistics in particular, centralized cloud-based EMT was envisioned in \cite{MFO, OVRP} to take advantage of similarities in the graph structures of vehicle routing problem (VRP) instances toward rapid optimization. The application of EMT to other forms of graph-based optimization tasks with potential use in manufacturing have also been explored in \cite{thang2021adaptive, dinh2020multifactorial}.

Despite some success, there are still challenges in reliably implementing EMT for combinatorial optimization tasks ubiquitous in manufacturing and operations research. A key issue is that of solution representational mismatch, which can lead to \emph{negative transfer} \cite{ORs}. For instance, consider unifying two VRPs in EMT that are defined using different node labels/indices even though their underlying customer distributions are similar. Due to the resultant label mismatch, genetic transfers under standard permutation-based solution representations would lead to suboptimal (or even confounding) exchange of routes or subroutes between tasks.

Two recent research avenues hold promise in overcoming the aforementioned challenge. The first entails departure from the usual direct transfer of solution prototypes in EMT. Instead, the \emph{transfer of higher-order solution construction heuristics} that are agnostic to low-level solution representations is proposed (as a form of multitask hyper-heuristic); both heuristic selection \cite{heuristic} and generative approaches \cite{zhang2021multitask} have been put forward, showing greater robustness to representational mismatches in EMT. The second research avenue deals with \emph{learning solution representations}, transforming problem instances in a manner that minimizes inter-task representational mismatch. An illustration of this idea is depicted in Fig. \ref{Fig:CVRP}, where two VRP instances ($VRP_1$ and $VRP_2$) with seemingly dissimilar customer distributions and node labelling are examined. However, through an \emph{isometric} transformation (comprising rotation and translation operations) of the nodes in $VRP_2$ (which preserves shortest routes), a new representation scheme that better aligns both tasks is obtained \cite{solre}.

\begin{spacing}{1.5}
\noindent \emph{$\bullet$ Case study in last-mile logistics planning \cite{EMT-CVRP}}
\end{spacing}

\begin{figure*}
  \centering
  \subfigure[\{$PDP_1$, $PDP_2$\}]{
  \label{CVRP-cvg1}
  \includegraphics[width=0.24\textwidth]{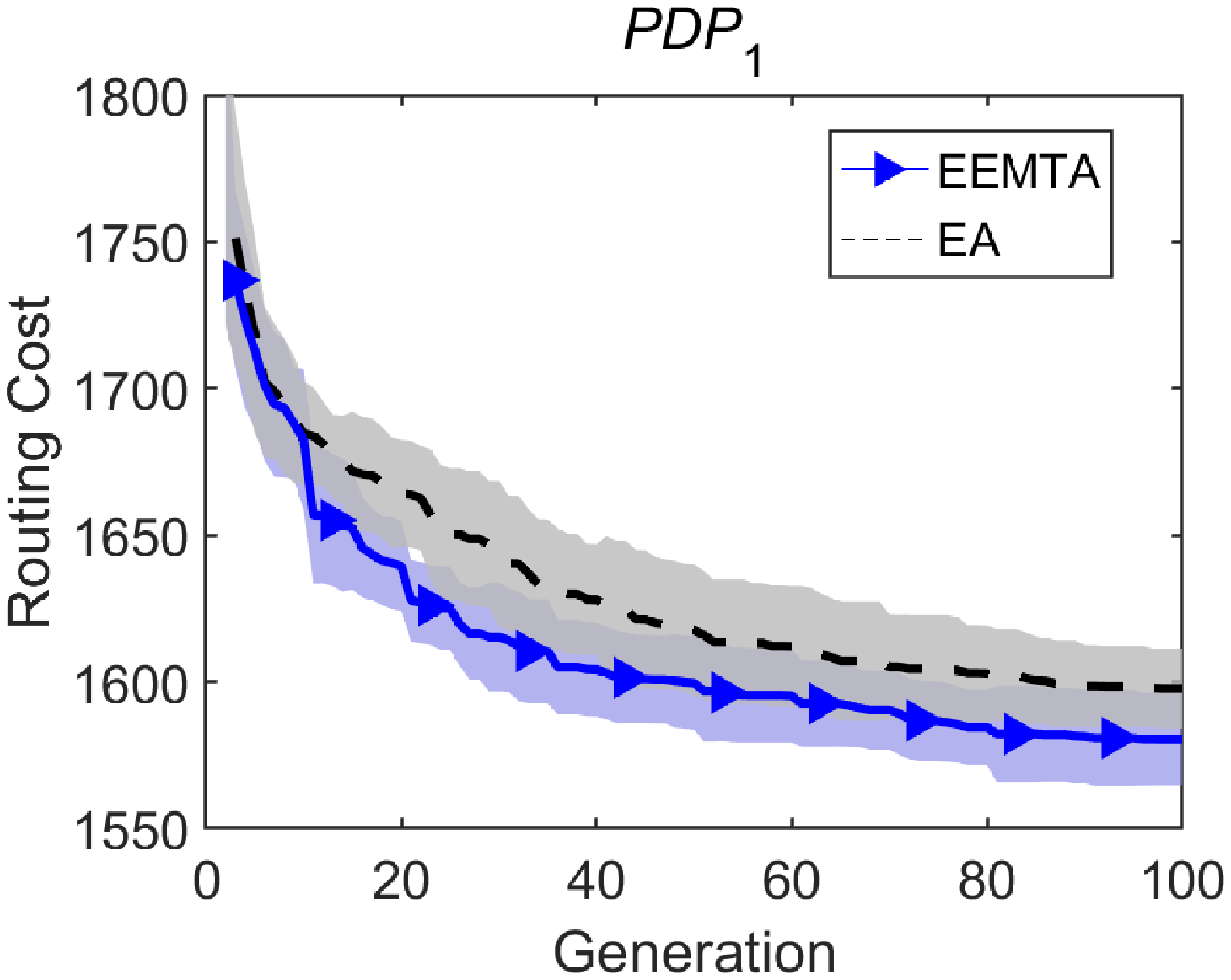}
  \includegraphics[width=0.24\textwidth]{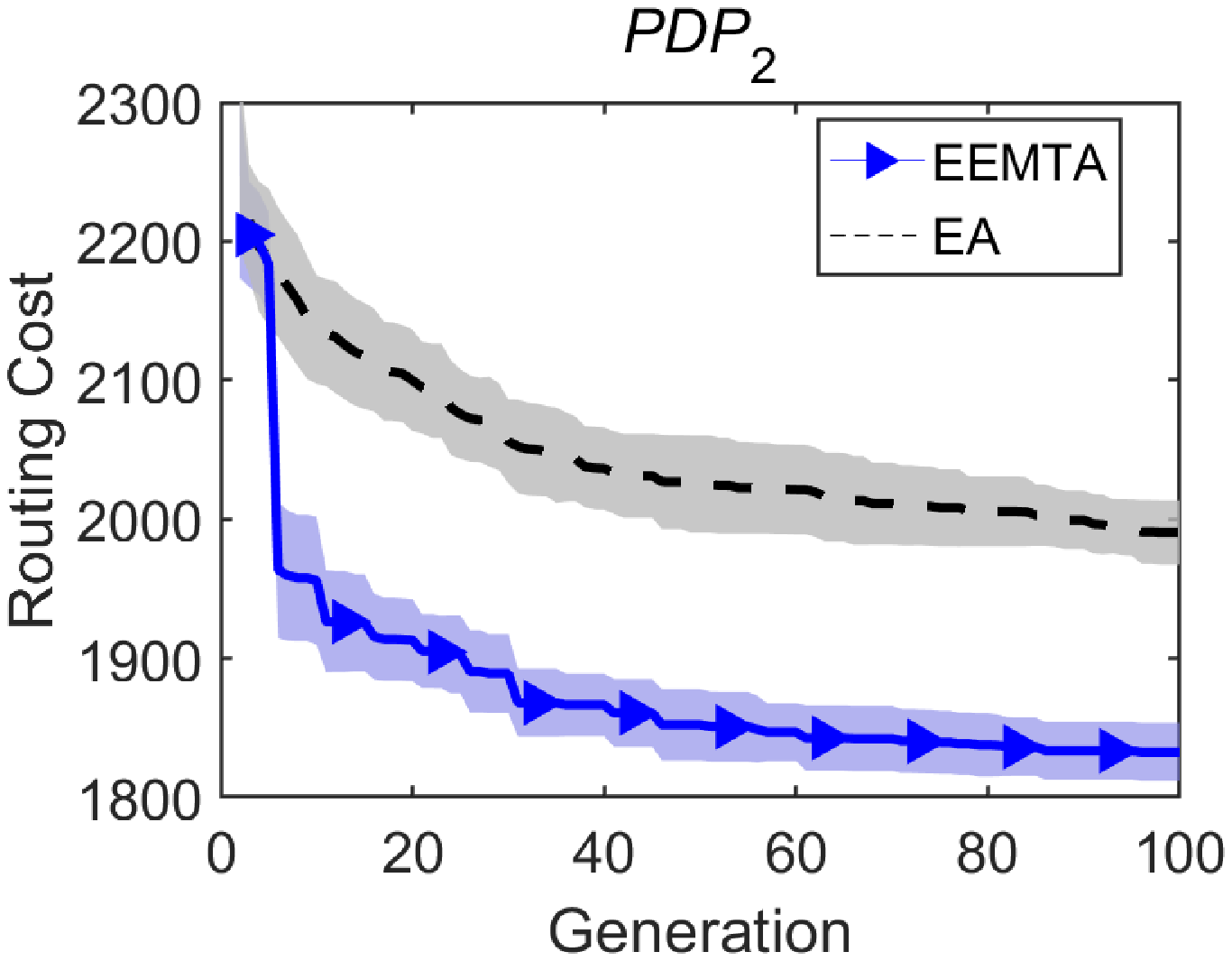}}
  \subfigure[\{$PDP_3$, $PDP_4$\}]{
  \label{CVRP-cvg2}
  \includegraphics[width=0.24\textwidth]{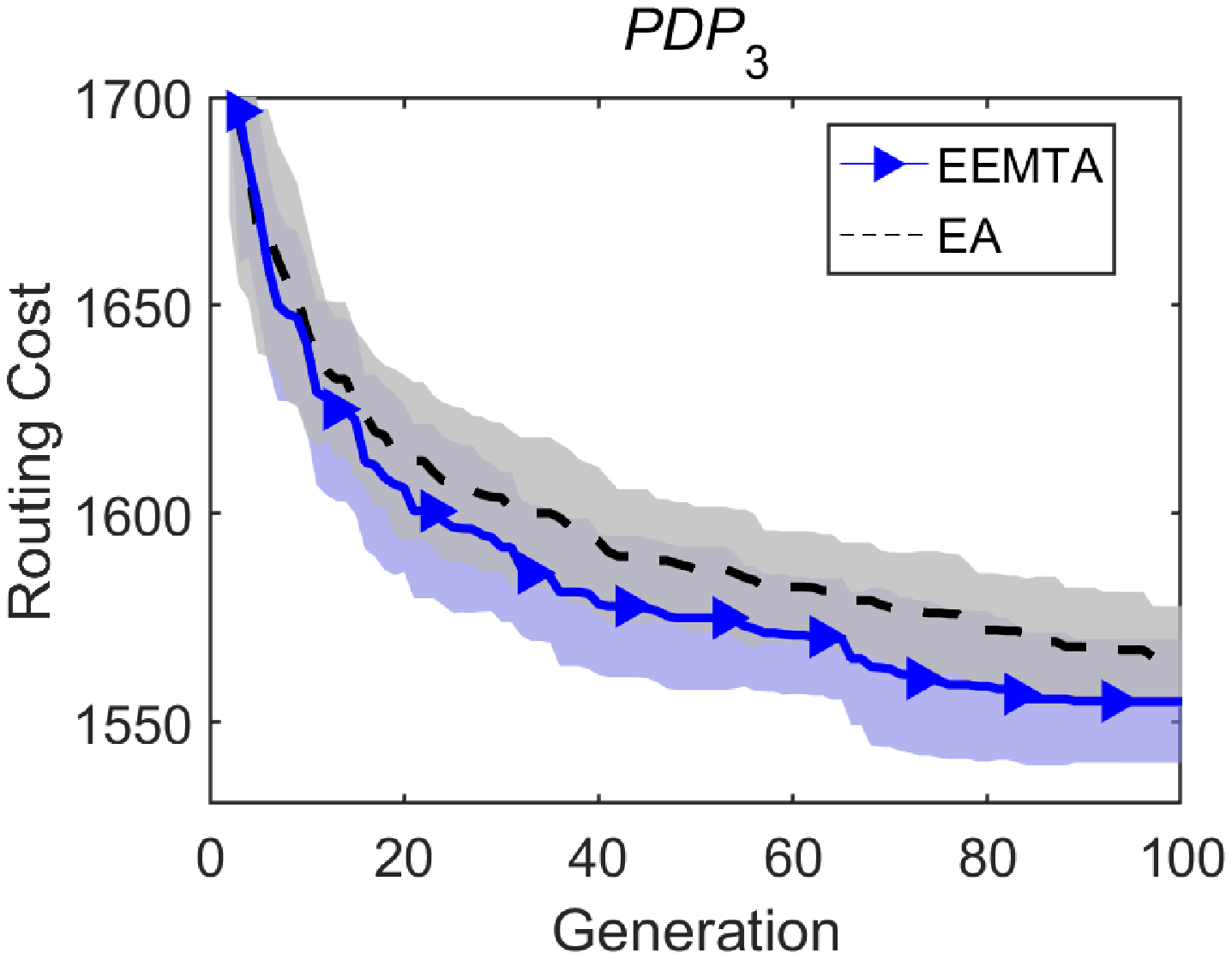}
  \includegraphics[width=0.24\textwidth]{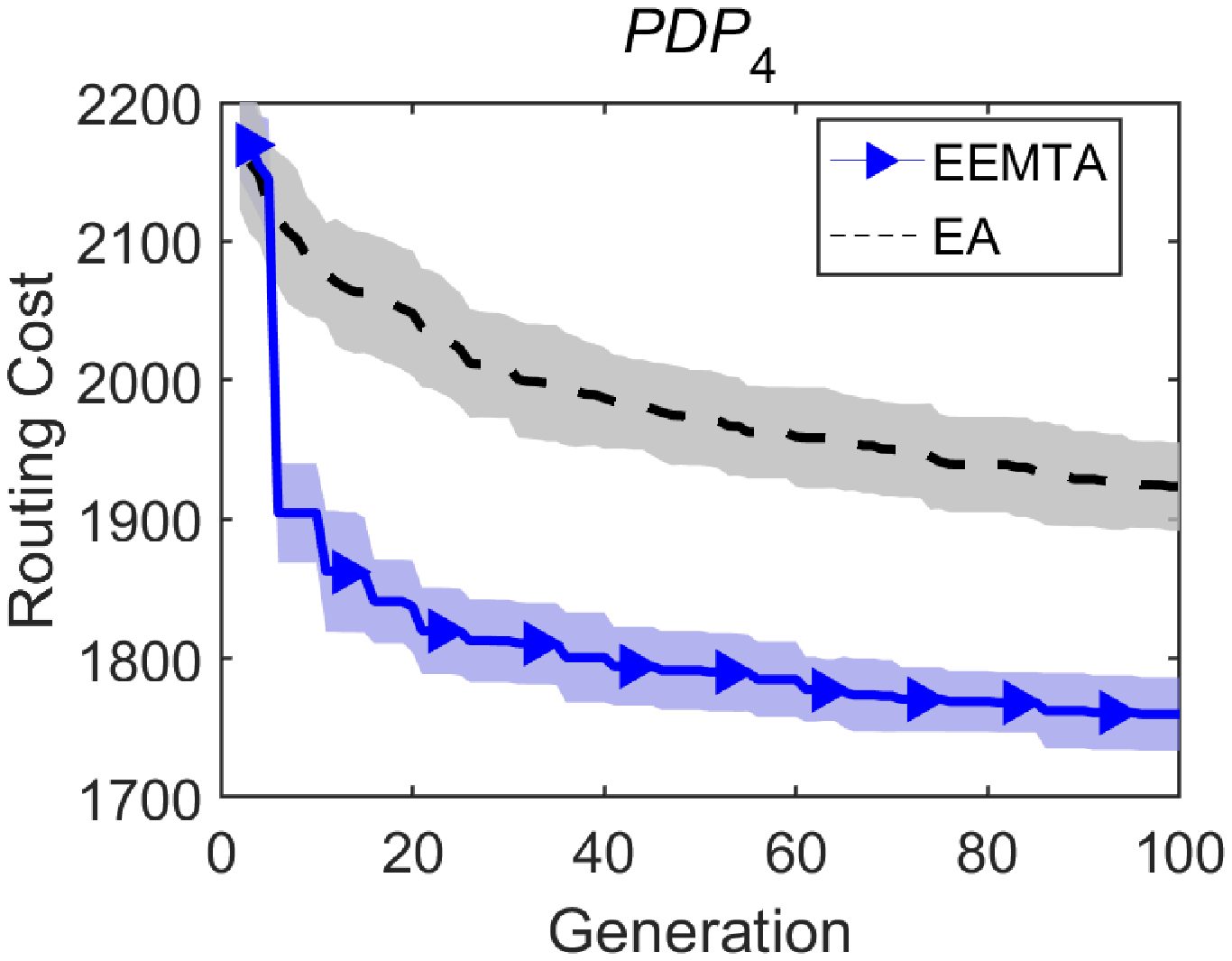}}
  \caption{Convergence trends (in routing cost minimization) of the representation learning-based EEMTA and a single-task EA on (a) \{$PDP_1$, $PDP_2$\} and (b) \{$PDP_3$, $PDP_4$\}. Results are obtained from \cite{EMT-CVRP}. The shaded area spans 1 standard deviation on either side of the mean performance.}
  \label{Fig:CVRPcvg}
\end{figure*}

\noindent Following on from the discussions above, a case study on real-world package delivery problem (PDP) instances \cite{EMT-CVRP} from a courier company in Beijing, China, is presented. The PDP is a variant of the NP-hard VRP, where the objective function pertains to minimizing total routing costs in servicing a set of geographically distributed customers (as illustrated in Fig. \ref{Fig:CVRP}) with a fleet of capacity constrained vehicles located at a single or multiple depots. The results presented hereafter are for an \emph{explicit} EMT combinatorial optimization algorithm (EEMTA for short) whose uniqueness lies in incorporating solution representation learning via sparse matrix transformations to facilitate the transfer of useful information across tasks. The reader is referred to \cite{EMT-CVRP} for full details of the EEMTA and the algorithmic settings used in the experimental study.

The experiments were conducted on four PDP requests that were paired to form two examples of MTO. The pairing was done based on customer distributions, with the resulting MTO formulations referred to as \{$PDP_1$, $PDP_2$\} and \{$PDP_3$, $PDP_4$\}, respectively. The convergence trends achieved by the EEMTA and the baseline single-task EA (hybridized with local search heuristics) are presented in Fig. \ref{Fig:CVRPcvg}. As can be seen in the figure, the EEMTA was able to achieve some extent of performance speed up across all four tasks. Multitasking provided an impetus to the overall search, whilst strongly boosting outcomes of the initial stages of evolution on $PDP_2$ and $PDP_4$ in particular.

\subsection{Category 6: EMT in Software and Services Computing}

\begin{figure*}
  \centering
  \includegraphics[width=0.8\textwidth]{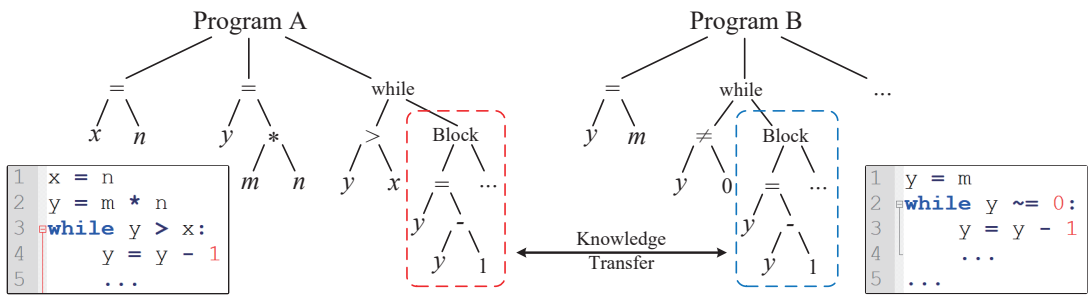}
  \caption{Two programs A and B concerning different tasks but with similar abstract syntax tree representations are depicted. Knowledge encoded in common subtrees could be efficiently transferred and reused through EMT to enhance the performance of an imagined automated program generator.}
  \label{Fig:Code}
\end{figure*}

\noindent Many problems in software engineering can eventually be converted into \emph{optimization} problem instances. Examples include finding the minimum number of test cases to cover the branches of a program, or finding a set of requirements that would minimize software development cost while ensuring customer satisfaction. The objective functions of such tasks generally lack a closed form, hence creating a niche for black-box search methods like EAs---underpinning the field of \emph{search-based software engineering} \cite{harman2012search}. What's more, as software services increasingly move to public clouds that simultaneously cater to multiple distributed users worldwide, a playing field uniquely suited to EMT emerges. A schematic of EMT's potential in this regard is highlighted in Fig. \ref{Fig:Code}, where the scope of joint construction/evolution of two distinct programs by the efficient transfer and reuse of common building-blocks of code is depicted.

Concrete realizations of this idea for web service composition (WSC) have been studied in the literature \cite{CCSC-1, CCSC-2}. The composition was achieved in \cite{CCSC-2} by formulating the problem as one of permutation-based optimization, where solutions encode the coupling of web services into execution workflows. Given the occurrence of multiple similar composition requests, a joint MTO formulation was proposed. The experiments compared three permutation-based variants of the MFEA against a state-of-the-art single-task EA on popular WSC benchmarks. The results showed that multitasking required significantly less execution time than its single-task counterpart, while also achieving competitive (and sometimes better) solution quality in terms of quality of semantic matchmaking and quality of service.

In what follows, we delve into a specific use-case in software testing that precisely fits the MTO problem setting with a set of objective functions and a set of corresponding solutions being sought.

\begin{spacing}{1.5}
\noindent \emph{$\bullet$ Case study in software test data generation \cite{Branch}}
\end{spacing}

\noindent In \cite{Branch}, the ability of EMT to guide the search in software branch testing by exploiting inter-branch information was explored. Each task in MTO represented a \emph{branch} of a given computer program, with the objective of finding an input such that the control flow on program execution (resulting from that input) would bring about the branch. Successfully achieving this is referred to as branch \emph{coverage}. Hence, the overall problem statement, given multiple branches, was to find a set of test inputs that would maximize the number of branches covered. (Optimal coverage could be less than 100\% since certain branches could be \emph{infeasible}, and hence never covered.)

In the experimental study, 10 numerical calculus functions written in C, extracted from the book \emph{Numerical
Recipes in C: The Art of Scientific Computing} \cite{press1988numerical}, were considered. The inputs to these functions are of integer or real type. Two EMT algorithm variants (labelled as MTEC-one and MTEC-all, indicating the number of tasks each candidate solution in a population is evaluated for) that seek to jointly cover all branches of a program were compared against a single-task EA tackling each branch independently. Table \ref{Table-BT} contains the averaged coverage percentage obtained by all algorithms over 20 independent runs, under uniform computational budget. The table reveals that MTEC, by virtue of leveraging inter-task information transfers, achieved competitive or superior coverage performance than the independent search approach on the majority of programs.

\begin{lrbox}{\tablebox}%
\begin{tabular}{c c | c c c }
  \hline
  \hline
  Program & Branches & MTEC-one & MTEC-all & Single-task EA \\
  \hline
  plgndr & 20 & \textbf{100} & \textbf{100} & 99.58 \\
  \hline
  gaussj & 42 & \textbf{97.62} & \textbf{97.62}  & \textbf{97.62}\\
  \hline
  toeplz & 20 & \textbf{85} & \textbf{85} & 84.75 \\
  \hline
  bessj & 18 & \textbf{100} & \textbf{100} & \textbf{100}\\
  \hline
  bnldev & 26 & \textbf{80.77} & \textbf{80.77} & 76.92 \\
  \hline
  des & 16 & \textbf{93.44} & 91.88 & \textbf{93.44} \\
  \hline
  fit & 18 & \textbf{97.5} & \textbf{97.5} & 92.78\\
  \hline
  laguer & 16 & \textbf{91.25} & \textbf{90.94} & 85 \\
  \hline
  sparse & 30 & 81.33 & \textbf{90} & 88\\
  \hline
  adi & 44 & \textbf{59.09} & \textbf{59.09} & 56.25 \\
  \hline
  \hline
\end{tabular}
\end{lrbox}
\begin{table}
\centering
\small \caption{The coverage percentage obtained by MTEC-one, MTEC-all and single-task EA over 20 independent runs. Best values are marked in bold. Reported results are obtained from \cite{Branch}.}\label{Table-BT} \scalebox{0.9}{\usebox{\tablebox}}
\end{table}

\section{Forging a Future for EMT in Multi-\emph{X} Problems} \label{S4}

\noindent The discussions heretofore provided an overview of EMT methodologies, and examples where these methods have been explored in real-world contexts. A representative set of applications from the literature were summarized in half-dozen broad categories, spanning topics in data science, complex engineering design, operations research, etc., offering a bird's eye view of the potential impact of EMT. This section looks toward the future of the field. An admittedly inexhaustive list of recipes is presented by which problem formulations of general interest---cutting across different application domains---could be newly cast in the light of EMT. The formulations fall under the umbrella of multi-\emph{X} EC \cite{MultiX}, since they stand to gain from the implicit parallelism of EAs in sampling, evaluating and processing multiple candidate solutions at once. It is hoped that our discussions will spark future efforts on pushing the envelope of population-based search further with both implicit and explicit EMT strategies.

\subsection{EMT in Multi-Objective, Multi-Constrained Problems}
\noindent Over recent decades, the solving of multi-objective optimization problems (MOPs) has greatly benefited from the capacity of EAs to generate approximations to a full Pareto set in a single run \cite{deb2002fast}. The universality of MOPs in decision-making has thus opened the door for EAs to wide-ranging disciplines. However, it has been shown that as the number of objective functions increases (referred to as \emph{many}-objective optimization problems, or MaOPs for short), the convergence rate of EAs may begin to slow down due to severe weakening of selection pressures \cite{ishibuchi2008evolutionary}. It is to remedy this shortcoming that motivates the revisit of MaOPs through the lens of EMT.

\emph{Lemma 1} of \cite{gupta2019blessing} suggests that an MaOP could be simplified into several MOPs, such that points on the Pareto front of an MOP also belong to the Pareto front of the target MaOP. Hence, the lemma naturally leads to a recipe for turning MaOPs into MTO problem formulations through the generation of a series of \emph{auxiliary} multi-objective optimization tasks. The known efficacy of EAs for MOPs could then be harnessed in an EMT algorithm to solve the main MaOP, with intuitive guarantees of useful inter-task information transfer. Notably, a different but associated idea has already been studied in \cite{largeEMT1}, where a large-scale MaOP is transformed into MTO and solved using the MFEA. The experimental results showed that, with limited computational budget, the multitask approach outperformed state-of-the-art baselines on benchmark MaOPs.

Similar to the recipe above, one can imagine that given a multi-constrained problem (or combined multi-objective, multi-constrained problem), simplified auxiliary tasks may be generated by dropping out some of the constraints \cite{qiaoevolutionary}. As long as the a priori unknown \emph{active constraints} are preserved, it is likely that solutions evolved for the auxiliary tasks would transfer beneficially to the main task at hand.

\subsection{EMT in Multi-Fidelity Optimization}
\noindent Multi-fidelity optimization is arguably a precise fit for MTO, and, by extension, for EMT. A population of candidate solutions is evolved to solve lower-fidelity tasks (with less accurate but cheap function evaluations) jointly with the high-fidelity (accurate but expensive) target problem instance---with the goal of reducing the load on high-fidelity analysis. The lower-fidelity tasks thus serve as catalysts to help quickly solve the target. Given \textit{K} tasks, where the $\textit{K}$-th is the target, the MTO can then be stated as:
\begin{equation} \label{Eq:multifidelity}
\begin{split}
   \{\mathbf{x}^*_1, \mathbf{x}^*_2, \dots, \mathbf{x}^*_{\textit{K}-1}, \mathbf{x}^*_{\textit{K}}\}  = \mathop{\arg\max} \{f^{low}_{1}(\mathbf{x}), f^{low}_{2}(\mathbf{x})\\
    \dots, f^{low}_{\textit{K}-1}(\mathbf{x}), f^{high}_{\textit{K}}(\mathbf{x})\},
\end{split}
\end{equation}
where the $f^{low}_{i}$'s represent the low-fidelity objective functions, and $f^{high}_{K}$ is their high-fidelity counterpart.

The setup of Eq. (\ref{Eq:multifidelity}) has widespread practical applicability. It has been alluded to previously in Section \ref{S3}, in connection with data science pipelines (for small to big data transfers) and safe UAV path planning. Engineering design also forms a major application area, where low-fidelity models extensively used for preliminary designs can be readily integrated into MTO frameworks. An illustrative case study was carried out in \cite{operational-indices}, where models with different levels of accuracy were combined in MTO for the multi-objective optimization of beneficiation processes; a variant of the MO-MFEA was utilized to this end. Multitasking across local and global models in surrogate-assisted optimization was considered in \cite{ExOP1}. Furthermore, a generalized EMT algorithm crafted for multi-fidelity problems in particular was proposed in \cite{GMFEA}.

\subsection{EMT in Multi-Level Optimization}
\noindent Multi-level optimization is characterized by mathematical programs whose constraints include a series of optimization problems to be solved in a predetermined sequence. For simplicity, our discussion here is limited to situations where only a single such constraint exists, forming what is typically referred to as a \emph{bilevel optimization} problem \cite{sinha2017review}. A sample formulation of a bilevel program is as follows:
\begin{equation} \label{Eq:BLOP}
\begin{split}
    \mathop{\min}_{\mathbf{x}_u \in \mathcal{X}_u}f_u(\mathbf{x}_u, \mathbf{x}_l^*), \;
    s.t. \; \mathbf{x}_l^* \in \mathop{\arg\max}_{\mathbf{x}_l \in \mathcal{X}_l}f_l(\mathbf{x}_u, \mathbf{x}_l),
\end{split}
\end{equation}
where $f_u$ is the upper-level objective function and $f_l$ is the lower-level objective function. The setup in Eq. (\ref{Eq:BLOP}) has manifold real-world applicability, with examples in environmental economics, optimal design, cybersecurity, and others \cite{sinha2017review}.

In the regime of black-box search, solving Eq. (\ref{Eq:BLOP}) may however give rise to computational bottlenecks in having to repeatedly optimize lower-level problem instances corresponding to different candidate solutions $\{\mathbf{x}_{u,1}, \mathbf{x}_{u,2}, \mathbf{x}_{u,3}, \dots\}$ at the upper level. It is in addressing this fundamental issue that EMT is expected to excel. By viewing the lower-level through the lens of EMT, a set of optimization tasks can be jointly solved as part of a single MTO setting as:
\begin{equation} \label{Eq:MT-BLOP}
   \mathbf{x}^*_{l,i} = \mathop{\arg\max}_{\mathbf{x}_l \in \mathcal{X}_l}f_{l}(\mathbf{x}_{u,i},\mathbf{x}_l), \text{for} \;\mathbf{x}_{u,i} = \{\mathbf{x}_{u,1}, \mathbf{x}_{u,2}, \dots\}.
\end{equation}
The recipe in Eq. (\ref{Eq:MT-BLOP}) was first explored in \cite{Bilevel}, under the intuitive assumption that similar upper-level candidate solutions would lead to lower-level problem instances amenable to inter-task transfers. An application optimizing the complete manufacturing cycle of lightweight composites substantiated this intuition, giving approximately 65\% saving in computation time compared to a standard evolutionary bilevel algorithm. In \cite{Minimax}, the authors considered solving expensive \emph{minimax} optimization---derived by setting $f_u = f_l$ in Eq. (\ref{Eq:BLOP})---via EMT. The resultant worst-case formulation was used to model a robust airfoil design problem, with experimental results showing that a surrogate-assisted MFEA vastly outperformed all the baseline algorithms. (It is contended that the success of \cite{Minimax} could be extended to multi-objective minimax problems \cite{eisenstadt2016novel, zychowski2018addressing} as well.)

\subsection{EMT in Multi-Scenario Optimization}
\noindent Imagine designing cars for various driving conditions, international markets (e.g., Asian, American), types of use (e.g., taxi, family car), or other \emph{scenarios}. During design optimization, every scenario could lead to different mathematical representations of the objective functions, even though their physical interpretations remain the same. For instance, let $S = \{1, 2, \dots, K\}$ be a set of scenarios, then a general formulation of a multi-scenario multi-objective optimization problem (MSMOP) may be stated as \cite{fadel2005multi, wiecek2007multi}:
\begin{equation} \label{Eq:MSMOP}
   \mathop{\max} \{\lbrack f_i^{1}(\mathbf{x}), f_i^{2}(\mathbf{x}), \dots, f_i^{m_i}(\mathbf{x})\rbrack, i \in S\},\; s.t.\; \mathbf{x} \in \mathcal{X}.
\end{equation}
Here, $m_i$ is the number of objectives in the $i$-th scenario, and $\mathcal{X}$ is a unified search space. A straightforward all-at-once approach tackles Eq. (\ref{Eq:MSMOP}) by fusing all the objective functions together into a gigantic MaOP. This may however lead to tractability issues and the return of solutions that do not belong to the Pareto set of individual scenarios. Hence, the solving of each scenario as a separate task was advocated in \cite{wiecek2007multi}, with post-hoc coordination between the tasks. Clearly, such a recipe for MSMOPs is ideally suited for the implicit parallelism and inter-task transfers of EMT, facilitating the discovery of solutions that are skilled for multiple scenarios.

A real-world study of such multi-scenario optimization was carried out in \cite{power-flow}, where EMT was used to support intra-hour optimal power flow under rapid load variations. Multiple scenarios were generated to accurately represent the variations in power demand, and the MFEA was used to derive optimized solutions for all scenarios in a proactive look-ahead manner. The obtained solution set could then be used as explicit setpoints to correctively control power generation---thus improving overall operational economy.

\section{Conclusions}\label{S5}
\noindent Evolutionary multitasking (EMT) is an emerging paradigm for jointly solving multiple tasks in a single optimization run. The basic idea is to allow tasks to exchange information, transferring evolved skills amongst one another to facilitate the efficient discovery of high-quality solutions. A wealth of research has been conducted in recent years to turn this idea into computational algorithms.

The main aim of this paper is to draw the attention of optimization practitioners to the many real-world use-cases of EMT algorithms. To this end, several application-oriented explorations of multitasking were reviewed in Section \ref{S3}. These were categorized into half a dozen broad categories, enabling readers to zoom in on applications of their choice. It is worth highlighting that even at the time of this article's review, a number of applied EMT works have appeared in the literature. Examples include an EMT method crafted for multigroup collision-avoidance control for UAV swarms \cite{luo2021grpavoid} (which outperformed state-of-the-art collision-avoidance models), a multiple surrogate assisted multitask algorithm for production optimization in petroleum engineering \cite{zhong2022surrogate}, among others \cite{wang2022evolutionary, ge2022mdde, shi2022multicriteria}.

Transcending specific application areas, Section \ref{S4} provided a set of recipes by which selected problem formulations of general interest could be recast as EMT instances. This inexhaustive list (which is expected to grow in the years to come) reveals tricks by which the concept of MTO can be uniquely leveraged, together with the implicit parallelism of EAs, to tackle difficult optimization problems.  \emph{The transformation of multi-\emph{X} optimization into MTO thus unpacks a new box of tools for practitioners, going beyond the typical rendition of EMT for simultaneously solving distinct problem instances}.

It must however be acknowledged that there are still fundamental challenges for EMT to reliably transition from concept to demonstration to deployment. Recent methodological reviews on EMT detail these issues \cite{survey1, wei2021review}. Within any application domain, the success of EMT depends on its ability to consistently deliver performance gains over state-of-the-art single-task algorithms/models. Ensuring \emph{scalability} to many-tasks and \emph{robustness} when faced with unrelated tasks is also vital to minimize/alleviate performance degradation. Hence, theoretical research on \emph{what, how, and when} to transfer in EMT remains paramount, in conjunction with the rigorous experimental comparison of new methods against competitive multi- \emph{and} single-task solvers in scenarios with related, unrelated, and a mix of related and unrelated tasks.

\section*{Acknowledgement}
Abhishek Gupta was supported by the A*STAR AI3 HTPO seed grant C211118016 on Upside-Down Multi-Objective Bayesian Optimization for Few-Shot Design. The work was also supported in part by the Cyber-Physical Production System Research Program, under the IAF-PP Grant A19C1a0018. Yaqing Hou was supported by the National Natural Science Foundation of China under Grant 61906032.

\bibliographystyle{IEEEtran}
\bibliography{ref}

\end{document}